\relax
\documentclass[letterpaper]{article} % DO NOT CHANGE THIS
\usepackage{aaai22}  % DO NOT CHANGE THIS
\usepackage{times}  % DO NOT CHANGE THIS
\usepackage{helvet}  % DO NOT CHANGE THIS
\usepackage{courier}  % DO NOT CHANGE THIS
\usepackage[hyphens]{url}  % DO NOT CHANGE THIS
\usepackage{graphicx} % DO NOT CHANGE THIS
\usepackage{natbib}  % DO NOT CHANGE THIS AND DO NOT ADD ANY OPTIONS TO IT
\usepackage{caption} % DO NOT CHANGE THIS AND DO NOT ADD ANY OPTIONS TO IT
\DeclareCaptionStyle{ruled}{labelfont=normalfont,labelsep=colon,strut=off} % DO NOT CHANGE THIS
\usepackage{algorithm}

\usepackage{booktabs}

\newcommand{\red}[1]{\textcolor{red}{#1}}

\usepackage{amssymb}
\usepackage[noend]{algpseudocode}
\usepackage[table]{xcolor}
\usepackage{multirow}
\usepackage{mathtools}
\usepackage{comment}
\newcommand{\green}[1]{\textcolor{green}{#1}}
\newcommand{\gray}[1]{\textcolor{gray}{#1}}
\def\etal{\emph{et al}.}
\def\eg{\emph{e.g}.}
\def\ie{\emph{i.e}.}

\title{Towards To-a-T Spatio-Temporal Focus for Skeleton-Based Action Recognition}

\author {
    Lipeng Ke\textsuperscript{\rm 1},
    Kuan-Chuan Peng\textsuperscript{\rm 2},
    Siwei Lyu\textsuperscript{\rm 1}
}
\affiliations {
    \textsuperscript{\rm 1} University at Buffalo, State University of New York\\
    \textsuperscript{\rm 2} Mitsubishi Electric Research Laboratories\\
    lipengke@buffalo.edu, kpeng@merl.com, siweilyu@buffalo.edu
}

\usepackage{bibentry}
\begin{document}

\maketitle

\begin{abstract}
Graph Convolutional Networks (GCNs) have been widely used to model the high-order dynamic dependencies for skeleton-based action recognition.
% Most existing approaches do not model joints' learnable dynamic connection topology or intensity,
Most existing approaches do not explicitly embed the high-order spatio-temporal importance to joints' spatial connection topology and intensity, and they do not have direct objectives on their attention module to jointly learn when and where to focus on in the action sequence. 
% on the adjacency matrices in GCNs to model the joints' spatial connection topology and intensity, 
%and they do not have direct objectives enforced on the adjacency matrices to ensure the consistency of spatio-temporal focus between adjacency matrices and the classifier
% and they do not have direct objectives to jointly enforce the attention module where and when to focus in the sequence. %based on the spatio-temporal attention.
To address these problems, we propose the To-a-T Spatio-Temporal Focus (STF), a skeleton-based action recognition framework that utilizes the spatio-temporal gradient to focus on relevant spatio-temporal features. We first propose the STF modules with learnable gradient-enforced and instance-dependent adjacency matrices to model the high-order spatio-temporal dynamics. Second, we propose three loss terms defined on the gradient-based spatio-temporal focus to explicitly guide the classifier when and where to look at, distinguish confusing classes, and optimize the stacked STF modules.
% we propose the STF exploration loss to enforce the GCN's focus to look over all the critical joints (spatial) and frames (temporal). We further design the STF divergence loss to enforce the GCN to focus on the discriminative spatio-temporal information to distinguish confusing classes. In addition, we propose the STF coherence loss to distillate the high-level module's focus to low-level modules.
STF outperforms the state-of-the-art methods on the NTU RGB+D 60, NTU RGB+D 120, and Kinetics Skeleton 400 datasets in all 15 settings over different views, subjects, setups, and input modalities, and STF also shows better accuracy on scarce data and dataset shifting settings. 
%We first propose the STF modules with learnable instance-dependent adjacency matrices to model the high-order dynamic graph, then use gradient-based spatio-temporal attention to guide the learning of STF.
%a skeleton-based action recognition framework that utilizes spatio-temporal gradient to focus To-a-T spatio-temporal features.
%We first propose the STF modules with learnable gradient-enforced and instance-dependent adjacency matrices to model the high-order spatio-temporal dynamics connections.
\end{abstract}

\section{Introduction}
\label{introduction}

%\kc{MC: missing citations.}
%\kc{The name of our proposed method (GAAMS) is not even introduced in Sec. 1. Please fix it.}

As a fundamental task in computer vision, action recognition \cite{simonyan2014two, varol2017long,carreira2017quo,si2018skeleton, Shi2019TwoStreamAG,Si_2019_CVPR,shi2020decoupled}
% Hu_2019_ICME,Yang_2020_IA,cho2020self,xiaolu_SPML19,Fan_2020_IA,Xie2018Memory,Liu_2017_CVPR, Song_2017_AAAI} 
has a wide range of applications in human-computer interaction \cite{liu2017enhanced}, video surveillance \cite{ji20123d}, and sports analysis \cite{herath2017going}. Existing action recognition methods can be categorized into video-based \cite{simonyan2014two}
% varol2017long,carreira2017quo} 
and skeleton-based \cite{si2018skeleton}
% Shi2019TwoStreamAG,Si_2019_CVPR,shi2020decoupled, Liu_2017_CVPR, Song_2017_AAAI}
which take the video and skeleton sequences as inputs, respectively. In recent years, with the improvement in hardware (\eg, MS Kinect) and skeleton extraction algorithms such as \cite{OpenPose}, the skeleton-based action recognition methods have received more attention for their low dimensional representation and robustness to the background changes \cite{johansson1973visual}. 

\begin{figure}[t]
\centering
\includegraphics[width=1\linewidth]{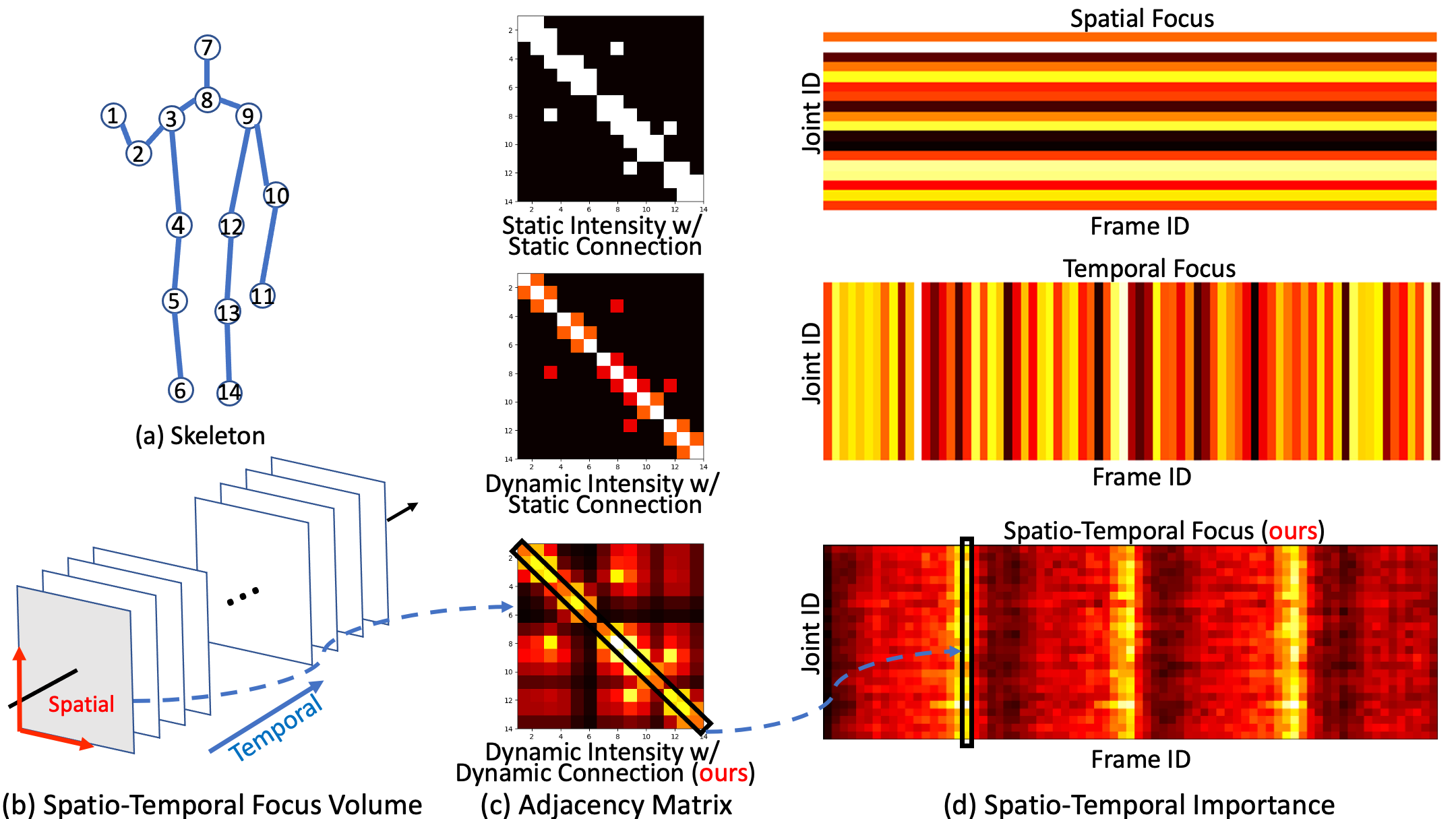}
\vspace{-1em}
\caption{Spatio-temporal focus: (a) is the layout of joints with their ID numbers, (b) \textbf{Spatio-temporal focus volume} stacks spatial focus along temporal dimension (c) \textbf{Adjacency matrix} models the joints connection topology and connection intensity, the top two rows either model static connection topology or static connection intensity; the bottom row models the dynamic connection topology and dynamic connection intensity; ours further guides the learnable dynamic adjacency matrix with spatio-temporal focus. (d) \textbf{Spatio-temproal focus} indicates when (temporal) and where (spatial) to emphasize. The top two rows model the spatial and temporal focus \textit{separately}, the bottom row (ours) models the spatio-temporal focus \textit{jointly}. 
%Spatio-temporal focus and GCN adjacency matrix. Spatial (a) and temporal (b) focus is the importance of joints within and across frames, respectively. (c) shows the difference between spatial, temporal, and spatio-temporal attentions, where the x/y axis is the temporal/spatial dimension. (d) is the skeleton layout with joint IDs. (e) is the binary adjacency matrix corresponds to the joint connections in (d). (f) has the same topology as (e) but has real-valued connection intensity. (g) is the proposed instance-dependent STF adjacency matrix that is not limited by the topology in (d).
% \caption{Our proposed method, spatio-temporal attention guidance (STF), uses spatio-temporal attention objectives to directly enforce the classifier to pay attention to all the critical and discriminative joints. Since we use spatio-temporal attention, our classifier pays attention to both the joint connection in spatial domain and their movements in temporal domain. STF encourages the instance-dependent adjacency matrix to be consistent with the spatio-temporal attention computed from the gradient which supports the classifier's prediction.
%We propose the attention guided adjacency matrix (AGAM) and encourage it to be consistent with the spatio-temporal attention computed from the gradient of the classifier which supports the classifier's prediction. \kc{Caption to be modified.}
}
\vspace{-.5em}
\label{fig:teaser}
\end{figure}

\begin{table}[t]
%\scriptsize
\footnotesize
\centering
\resizebox{1\columnwidth}{!}{
\begin{tabular}{@{}l@{\hspace{.3em}}c@{\hspace{.5em}}c@{\hspace{.5em}}c@{\hspace{1.2em}}c}
\toprule
%characteristics $\backslash$ methods & ST-GCN\cite{Yan2018SpatialTG} & AS-GCN\cite{li2019actional} &2s-AGCN~\cite{Shi2019TwoStreamAG} &GCN-NAS~\cite{Peng2020LearningGC} &MS-AAGCN~\cite{Shi2019SkeletonBasedAR} &MS-G3D\cite{liu2020disentangling}  &STF\\
%characteristic / method (group) &$G_1$ &$G_2$ &$G_3$ &\cellcolor{blue!30}STF\\
property / method (group) &$G_1$ &$G_2$ &$G_3$ &\cellcolor{blue!30}STF\\
\midrule
%\textbf{with} learnable adjacency matrix
% has \textbf{learnable} adjacency matrix &\green{Y} &\green{Y} &\red{N}  &\cellcolor{blue!30}\green{Y}\\
% %\textbf{with} dynamic topology of adjacency matirx
% adjacency matrix has \textbf{dynamic topology} &\green{Y} & \red{N}&\red{N}&\cellcolor{blue!30}\green{Y}\\
% %\textbf{with} loss directly supervising adjacency matrix
% has \textbf{loss} directly supervising \textbf{adjacency matrix} &\red{N} &\red{N}&\red{N} &\cellcolor{blue!30}\green{Y}\\
% %\textbf{with} attention loss to guide when and where to look at
% has \textbf{loss} directly supervising \textbf{attention} (modules) &\red{N} &\red{N} &\red{N} &\cellcolor{blue!30}\green{Y}\\
% \midrule
% conjunction of all the above &\red{N} &\red{N} &\red{N} &\cellcolor{blue!30}\textbf{\green{Y}}\\
adjacency matrix w/ \textbf{learnable dynamic intensity} &\green{Y} &\green{Y} &\red{N}  &\cellcolor{blue!30}\green{Y}\\
%\textbf{with} dynamic topology of adjacency matirx
adjacency matrix w/ \textbf{learnable dynamic topology} &\green{Y} & \red{N}&\red{N}&\cellcolor{blue!30}\green{Y}\\
%\textbf{with} loss directly supervising adjacency matrix
adjacency matrix w/ \textbf{spatio-temporal focus} &\red{N} &\red{N}&\red{N} &\cellcolor{blue!30}\green{Y}\\
%\textbf{with} attention loss to guide when and where to look at
attention (modules) w/ \textbf{direct supervision} &\red{N} &\red{N} &\red{N} &\cellcolor{blue!30}\green{Y}\\
\midrule
conjunction of all the above &\red{N} &\red{N} &\red{N} &\cellcolor{blue!30}\textbf{\green{Y}}\\
\bottomrule
\end{tabular}
}
\vspace{-.4em}
\caption{Comparison between STF and recent methods of skeleton-based action recognition. Among all the listed methods, STF models one of the most generalized types of adjacency matrices with explicit backward-knowledge spatio-temporal focus regularization. Method groups: $G_1$:~\cite{ye2020dynamic, shi2021adasgn,chen2021channel,zeng2021learning}, $G_2$:~\cite{song2020stronger,song2020richly,yang2020improving,li2020temporal,Cheng_ECCV20,Heidari2020stbln,Heidari2020pstgcn,liu2020disentangling,Yan2018SpatialTG,yang2020centrality,xiaolu_SPML19,Fan_2020_IA,Yang_2020_IA,li2019actional,qin2021fusing,chen2021learning,plizzari2020spatial,shi2020decoupled,Shi2019SkeletonBasedAR,Shi2019TwoStreamAG}, $G_3$:~\cite{li2020multi,pan2021spatio,cho2020self,Peng2020LearningGC,si2018skeleton,Liu_2017_CVPR,Song2017AnES,Xie2018Memory,huang2020spatio,Si_2019_CVPR}. 
}
\label{table:compare_related_works}
\vspace{-.7em}
\end{table}

Since the action sequence is a time series of human joint locations, we can represent it as a three-dimensional tensor, with the joints layout being the \textbf{spatial dimension} and joints movement in time series as the \textbf{temporal dimension}. To model the \textbf{temporal} information, 
earlier deep neural network based action recognition methods (\eg, \cite{Liu_2017_CVPR})  model the \textit{temporal} movement of joints across frames directly using recurrent neural network with long short-term memory. However, these methods do not explicitly consider the \textit{spatial} dependencies among different joints (Figure~\ref{fig:teaser}(d), middle-row). 
% Recently, Graph Convolutional Networks (GCNs), has enjoyed a great success in action recognition by modeling the skeleton topology more precisely using a graph. 
%In parallel, many methods \cite{Shi2019TwoStreamAG,si2018skeleton} model the skeleton topology as graphs and apply graph convolutional networks (GCNs) to model the spatial dependencies among joints, as shown in Figure~\ref{fig:teaser}(e)(f).
Subsequently, many methods \cite{si2018skeleton,liu2020disentangling,Peng2020LearningGC,Shi2019TwoStreamAG,li2019actional} model the skeleton topology as graphs and use graph convolutional networks (GCNs) to model the \textbf{spatial} connections (topology and intensity) among joints (Figure~\ref{fig:teaser}(c), top row). The \textit{static} topological connections of the joints are captured by the adjacency matrix in Figure \ref{fig:teaser}(c) top row, and the connection intensity learned from data is shown in Figure \ref{fig:teaser}(c) middle row.  %\cite{li2019actional}.
%with ${0, 1}$ to denote the connection of joints, or have learned float value \cite{} to denote the connection intensity between joints.
Both types of methods (Figure \ref{fig:teaser}(c) top \& middle rows) model the fixed physically constrained topology connections that cannot be adapted for varying temporal dependencies in the input sequence.
Some recent works~\cite{ye2020dynamic,shi2021adasgn,chen2021channel,zeng2021learning} model dynamic connection topology from data, but the dynamic connection topology is 
%Moreover, existing adjacency matrices are either predefined or 
generated from a forward pass through the model without any objective directly regularizing the adjacency matrices using spatio-temporal focus. 
% direct a loss directly regularizing the adjacency matrices
% backward knowledge learned from the prediction error. %\lk{another sentence to summarize}
%without backward knowledge to enforce the classifier to learn from the prediction error.
%\kc{Figure \ref{fig:teaser}(f) is NOT fixed topology adjacency matrix. You haven't explained the drawback of \cite{Shi2019TwoStreamAG}.}
%, more over these adjacency matrix mainly focus on the spatial connection with limited consideration of temporal domain. 
% However, these methods mostly rely on predefined adjacency matrix   of skeleton topology, which model the intrinsic relationship between skeleton joints \cite{shi2020decoupled}, and may not capture the complex high-order dependencies of joints across the spatio-temporal domains. 

\textbf{Spatio-temporal modeling} 
% In addition to modeling the topology of skeleton joint connections, when (temporal) and where (spatial) to focus on are another challenge in GCN-based action recognition methods. 
Since not all joints and frames are equally important for the recognition~\cite{shi2021adasgn}, and only specific joints (\textbf{spatial}) 
% in specific frames 
with specific motion (\textbf{temporal}) are critical to distinguish different action classes~\cite{xiaolu_SPML19}, finding these critical joints (spatial) and the motion patterns (temporal) \textit{jointly} in skeleton sequences is important for action recognition. However, most existing methods \cite{shi2020decoupled,Si_2019_CVPR,cho2020self,Xie2018Memory} simply create the attention modules using trainable parameters, which do not have the objectives to directly enforce the modules to capture the varying spatial and temporal patterns jointly.
%which do not have the objectives to directly enforce the classifier to capture the varying spatial and temporal pattern.
%varying sparse spatial and temporal attention.
% \kc{From the previous sentence, it's not intuitive why the attention modules of existing methods don't capture such the varying spatio-temporal attention. What's the purpose of mentioning ``sparse?"}
%which do not have a loss to capture only and all critical joints and their temporal importance. %temporal features in the temporal dimension
%Thus there is no loss function to guide the classifier's to make prediction based on all the critical joints in all the critical frames, to pay attention to the difference in spatio-temporal domain for different classes, to have consistent spatio-temporal attention across the GCNs modules. 

To better guide the classifier about when and where to look at jointly, and model the learnable dynamic joints connection in spatio-temporal domain, we propose the {\em To-a-T Spatio-Temporal Focus} method (termed as STF), which uses the joint spatial and temporal information jointly (shown in Figure~\ref{fig:teaser}(c) bottom).
% To better model spatial and temporal dependencies jointly in the skeleton sequence, we propose a learnable and instance dependant {\em spatio-temporal attention guided} GCN (termed as STF) to capture high-order dynamic adjacency matrix, Figure \ref{fig:teaser}(g).
Specifically, we extract the spatio-temporal focus that strongly influences the recognition in training by projecting the backward gradient to the spatio-temporal domain. 
% Then, we propose the STF modules that are guided by the obtained spatio-temporal focus to generate instance-dependent adjacency matrices that reflect not only the high-order topology connections but also the spatio-temporal importance given the input skeleton sequence.
Then we propose the STF modules to generate dynamic instance-dependent adjacency matrices, and we use the obtained spatio-temporal focus to regularize the dynamic adjacency matrices, such that the matrices reflect not only the high-order topology connections but also the spatio-temporal importance given the input skeleton sequence.
%by projecting the backward error to spatio-temporal domain.
Besides, we also use the gradient-based spatio-temporal focus to encourage the classifier to better emphasize on the critical spatio-temporal inputs and features. To achieve these goals, we introduce three loss terms: (1) the STF exploration loss to enforce the classifier to make prediction over {\em all} the critical joints; (2) the STF divergence loss to minimize the similarity of the focus for different classes, and (3) the STF coherence loss to focus on consistent spatio-temporal features across the stacked STF modules. 
% We build STF with the STF exploration loss, STF divergence loss, and STF coherence loss. 
Our proposed STF method outperforms the state-of-the-art skeleton-based action recognition methods in all 15 settings over different views, subjects, setups, and input modalities on the NTU RGB+D 60, NTU RGB+D 120, and Kinetics Skeleton 400 datasets.

Our contributions are summarized below: %\kc{Please claim the contributions about data-shift and scarce data experiments.}
\begin{itemize}
    \item We propose the To-a-T Spatio-Temporal Focus framework (STF) as a flexible framework trained with spatio-temporal gradient for skeleton-based action recognition.
    %\item In STF, we design the novel STF module that generates a learnable and instance-dependent adjacency matrix to model the high-order connections in spatio-temporal domain, and propose a loss to incorporate spatio-temporal focus to guide the learning of the adjacency matrix.
    % \item We design STF adjacency matrix loss to incorporate spatio-temporal attention to guide the learning of adjacency matrix. \kc{Why do you ignore the novelty of the 3 other losses in the contribution paragraph?}
    \item We design the novel STF module that generates dynamic connection topology and intensity, and propose a loss to incorporate the spatio-temporal focus to regularize the spatio-temporal connection topology and  intensity.  
    
    % \red{classification driven attention}
    % \item To the best of our knowledge, this is the first skeleton-based action recognition method that incorporates attention guided adjacency matrix for supervised learning of joint dependency modeling instead of using handcrafted graph topology.
    \item We propose three loss terms defined on the gradient-based spatio-temporal focus to explicitly guide the classifier when and where to look at, distinguish confusing classes, and optimize the stacked STF modules.
    %classifier's capacity to focus on the crucial spatio-temporal features and utilize the later STFe's attention to improve the earlier STFe's learning in stacked GCNs.
    %more multiple joint correlations.
    \item Our proposed STF framework outperforms the SOTA not only in three benchmarks but also in the scarce data and dataset shifting settings. %Our proposed STF framework can generalize on scarce data, and it is robustness on dataset shifting. 
\end{itemize}

%\begin{enumerate}[wide, labelwidth=!, labelindent=0pt]
%\end{enumerate}

\section{Related Works}
\label{related_works}

% \kc{Please remember to include 2021 references.}

%organize the following refs and argue why ours is better than them: DM-3DCNN~\cite{Ruiz20173DCO}, ST-GCN~\cite{Yan2018SpatialTG}, SR-TSL~\cite{si2018skeleton}, AS-GCN~\cite{li2019actional}, TS-SAN~\cite{cho2020self}, MS-G3D~\cite{liu2020disentangling}, 2s-AGCN~\cite{Shi2019TwoStreamAG}, 2s-NLGCN~\cite{Shi2018NonLocalGC}, GCN-NAS~\cite{Peng2020LearningGC}, MS-AAGCN~\cite{Shi2019SkeletonBasedAR}, MS-G3D~\cite{liu2020disentangling}, and VA-Fusion (TPAMI'19)

\noindent{\bf Graph neural network for action recognition}.
%Graph neural networks (GNNs) are neural networks that can capture the complex dependencies in graphs using deep neural networks \cite{zhou2018graph}. %GNNs have been widely used for feature extraction for general graphs.
Graph convolutional networks (GCNs) \cite{zhou2018graph} are a type of GNNs that extend convolutional operations to the adjacency matrices of structured data types that can be modeled as graphs. GCNs have been adapted widely to a number of applications including graph classification/embeding/prediction, link prediction, node classification etc. The first work applying GCN to  skeleton-based action recognition is ST-GCN~\cite{Yan2018SpatialTG}, which constructs a spatio-temporal graph with the joints as the nodes and skeletal connectivity as the edges to model joint dependencies. However, ST-GCN only considers joints that are directly connected in the skeleton, which limits its representation capacity. Subsequently, multi-scale GCNs \cite{li2019actional,liu2020disentangling} are proposed to capture dependencies among joints that are not neighbors in the skeleton graph. These methods use higher order polynomials of the adjacency matrix to aggregate features from non-adjacent joints. 2s-GCN \cite{Shi2019TwoStreamAG} further adapts the adjacency matrix to model the \textbf{learnable dynamic intensity} of the joints connection using an embedding function. 2s-GCN also popularizes the use of multi-stream inputs, such as joint, bone, joint motion, bone motion, angular, etc., for skeleton-based action recognition \cite{Shi2019TwoStreamAG,Shi2019SkeletonBasedAR,qin2021fusing}. 
A common drawback of these works is that they use the same adjacency matrices for different inputs, and the adjacency matrices they use only model the joints' physical topology connection, which limits their adaptivity to largely different and dynamic actions. Recently, Dynamic-GCN~\cite{ye2020dynamic} introduces the Context-Encoding-GCN to learn skeleton topology automatically, but existing methods with \textbf{learnable dynamic topology}~\cite{ye2020dynamic, shi2021adasgn,chen2021channel,zeng2021learning} (Table~\ref{table:compare_related_works} \textit{G$_1$}) 
do not have objectives directly enforced on their adjacency matrices to ensure that the spatio-temporal focus is correct. Thus we design the STF module that models dynamic connection topology and intensity with spatio-temporal focus that is embedded into the dynamic adjacency matrix for explicit enforcement. 
%do not have loss functions directly applied to incorporate the spatio-temporal focus.

%Nevertheless, common to these existing works \cite{Yan2018SpatialTG,Shi2019TwoStreamAG,li2019actional,liu2020disentangling} either rely on the skeletal connectivity of joints, or use embedding to model the connectivity without any constrains. 

%Thus, we propose the STF adjacency matrix to model the connection in a learnable and instance-dependent manner, and use spatio-temporal attention to guide the learning of the STF adjacency matrix.

\noindent{\bf Attention mechanism for action recognition}.
In parallel with the GCN approaches, the attention mechanism has been widely used in skeleton-based action recognition \cite{cho2020self, xiaolu_SPML19, Fan_2020_IA, Xie2018Memory, Liu_2017_CVPR, Song_2017_AAAI} since it was introduced to this task by Liu \etal~\cite{Liu_2017_CVPR} and Song \etal~\cite{Song_2017_AAAI}. The attention scheme can selectively focus on discriminative sets of joints within each frame of the inputs, and it can also exert different levels of influence on the outputs of different frames. 
Xie \etal~\cite{Xie2018Memory} proposed the Temporal Attention Recalibration Module and Spatio-Temporal Convolution Module to re-calibrate the temporal and spatio-temporal attention across the frames, respectively. Fan \etal~\cite{Fan_2020_IA} further proposed a method consisting of a self-attention branch and a cross-attention branch to utilize the scenario context information. Ding \etal~\cite{xiaolu_SPML19} improved the spatio-temporal attention by using an end-to-end attention-enhanced recurrent GCN to pay different levels of attention to different temporal and spatial joints. To capture long-term spatio-temporal relationships, Cho \etal~\cite{cho2020self} proposed the Self-Attention Network to extract high-level semantics capturing long-range correlations.

%Instead of using the aforementioned attention mechanisms to capture spatio-temporal dependencies, Shi \etal~and Yang \etal~\cite{shi2020decoupled, Yang_2020_IA} use attention mechanisms to learn the dependencies between joints that are not physically connected to replace the fixed adjacency matrix. Yang \etal~\cite{Yang_2020_IA} is the first work to combine GCN and attention mechanism for skeleton-based action recognition. Their method is based on a variant of GCN (termed as pseudo GCN) with temporal and channel-wise attention to resolve the fixed adjacency matrix. Shi \etal~\cite{shi2020decoupled} further propose spatio-temporal attention decoupling, decoupled position encoding, and global spatial regularization to model spatio-temporal dependencies between joints without the need to know their positions or mutual connections.
%without the requirement of knowing their positions or mutual connections.

However, the aforementioned attention mechanisms have two common problems. They do not consider spatial and temporal dependencies jointly, and they only receive implicit supervision from the final classification loss. These issues limit the accuracy of the attention-based methods when they are applied to the sequences corresponding to complex actions. To address these problems, we propose three loss terms defined on the gradient-based spatio-temporal focus to explicitly guide the classifier when and where to look at, distinguish confusing classes, and optimize the stacked STF modules.

We summarize the major weaknesses of existing works in Table~\ref{table:compare_related_works} to highlight our motivation and novelty. Our STF is the only one with direct supervision to extract gradient-based spatio-temporal attention, and we use the spatio-temporal focus to regularize the adjacency matrix with learnable dynamic topology and intensity of joint connection.
% These issues significantly limit the accuracy of the attention-based methods when applied to sequences corresponding to complex actions.
%To illustrate our novelty comparing with existing works, we compare our STF with recent works in Table~\ref{table:compare_related_works}.

%either do not utilize the spatial and temporal attention jointly, or use self-attention that do not have explicit supervision designed to regularize the attention other than the classification loss. Thus, we proposed an gradient based spatio-temporal attention that capture the importance in spatial and temporal domain jointly.% We also propose three spatio-tempoal attention guided losses to enforce the classifier to focus on the most critical joints and their movements in temporal domain to distinguish the most confusing classes. Moreover, we propose STF adjacency matrix module to generate a instance-dependent and high-order relationship without predefined topology, and proposed a loss to supervise the learning of STF adjacency matrix module with our spatio-temporal attention guidance. 

%\ko{I suggest that we argue why the learned weighting of the adjacency matrix should be consistent with the attention volume.}
%\kc{You forget to mention that you also propose $L_{G_k}$ to supervise AGAM.} 

\section{Our Method --- To-a-T Spatio-Temporal Focus}
\label{method}

\begin{figure*}
\centering{
\includegraphics[width=.995\linewidth]{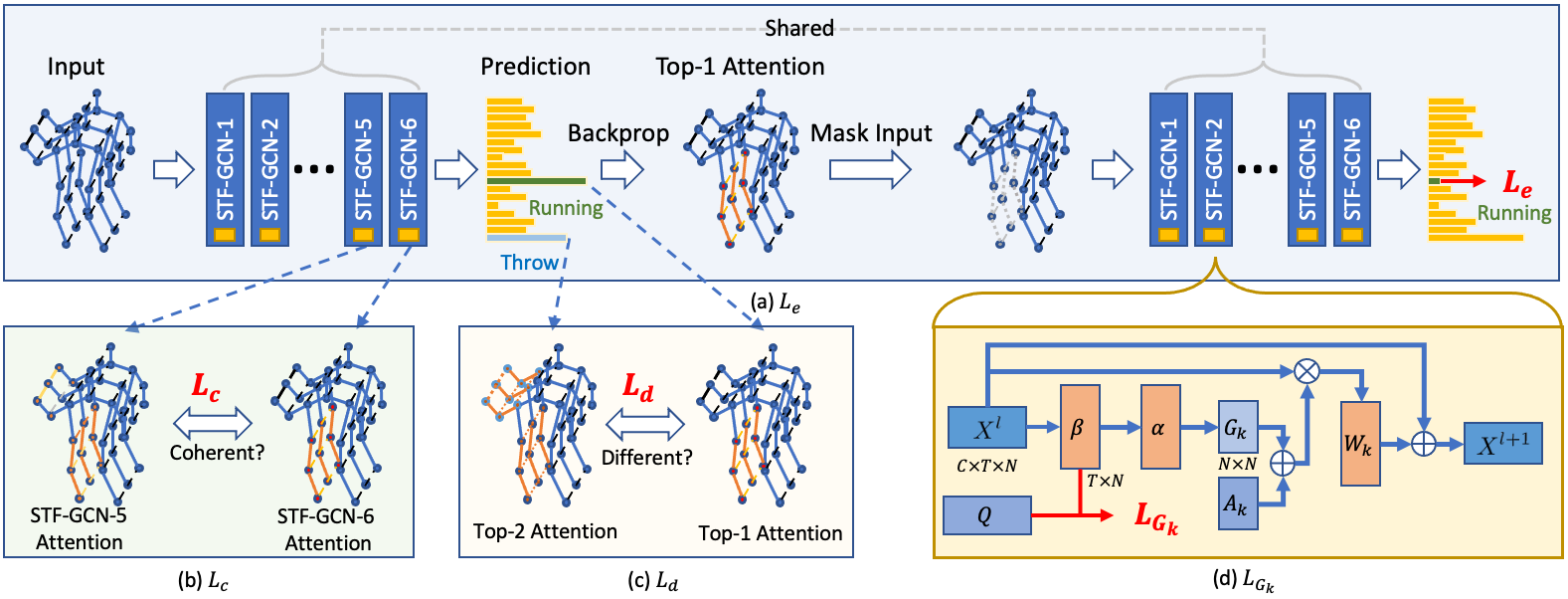}}
\vspace{-.2em}
\caption{The illustration of the proposed objectives (all the spatio-temporal focus is projected to the input sequence for visualization): (a) STF exploration loss $L_{e}$, which masks the input sequence according to the focused parts and minimizes the probability of the originally predicted class of the masked input sequence; (b) STF coherence loss $L_{c}$ enforces the focus to be coherent across the last two STF modules; (c) STF divergence loss $L_{d}$ enforces the top two predicted classes to have different focused parts; (d) STF adjacency matrix and its loss function $L_{G_k}$, which encourages the consistency between the spatio-temporal focus and STF adjacency matrix such that the adjacency matrix is adaptive, high-order and instance-dependent. The yellow boxes in STF-GCN-1$\sim$STF-GCN-6 are not shared (\ie, each STF module has its own STF adjacency matrix). 
}
%attention guided adjacency matrix AGAM and its supervision signal $L_{G_k}$ use the spatio-temporal attention to guide embedding to generate an adaptive, high-order and instance-dependent adjacency matrix.
\label{fig:pipeline}
\vspace{-.5em}
\end{figure*}

We propose a new framework, To-a-T Spatio-Temporal Focus (STF), for skeleton-based action recognition. STF utilizes the spatio-temporal focus acquired by backward gradients to improve the graph modeling in typical GCN and capture critical spatio-temporal features. We first propose the STF modules, which generate learnable and instance-dependent adjacency matrices. Then we use the gradient-based spatio-temporal focus to supervise the STF modules, such that the adjacency matrices capture both the high-order dynamic dependency and encode spatio-temporal importance. Second, we propose the STF exploration loss, STF divergence loss, and STF coherence loss to explicitly enforce the classifier to predict based on all critical joints and frames across the input, to focus on discriminative spatio-temporal features of confusing classes, and to have consistent spatio-temporal attention across the stacked STF modules, respectively. With these proposed losses, STF is better guided when and where to look at, distinguishing confusing classes, and distilling the high-level module’s focus to low-level modules. 

\subsection{Spatio-Temporal Focus (STF) Module}
\label{sec:gk}

%{\bf Notations.}
A human skeleton graph is defined as $(V, E)$,
where $V =\{v_1, ..., v_N \}$ is the set of $N$ nodes representing the joints, and $E$ is the edge set representing the bones by an adjacency matrix $A \in R^{N\times N}$.
The input skeleton sequence is formulated as a tensor $X \in R^{C\times T\times N}$, where $C$, $T$, and $N$ are the joint feature dimension (for 2D/3D input, $C$=2/3), number of frames, and number of joints, respectively.

In the vanilla adjacency matrix (Figure~\ref{fig:teaser}(c) top row),  $A_{i,j} = 1$ if the joints corresponding to nodes $v_i$ and $v_j$ are connected in the skeleton, and $0$ otherwise. $A$ is symmetric since the graph is undirected. We build up our STF module upon \cite{liu2020disentangling}, which encodes multi-scale adjacency matrices $A_k$. $A_k$ considers connections via fewer than $k$ intermediate joints as spatial connections at the scale of $k$.
%that is joints connected via less than $k$ intermediate joints will be considered as connected joints in $A_k$.
Since $A_k$ is still constrained by the natural topology of human skeleton, which cannot model the high-order relationship in the spatio-temporal domain, we propose an instance-dependent and learnable dynamic adjacency matrix $G_k$ (Figure~\ref{fig:teaser}(c) bottom row) to model the high-order dependency along with $A_k$. The features of the STF modules are computed as:
\begin{equation}
% X_t^{l+1} = \delta(\hat{D}^{-\frac{1}{2}}\hat{A}\hat{D}^{-\frac{1}{2}}X_t^{l}),
X^{l+1} = \sum_{k=1}^{K}W_kX^{l}(A_{k}+G_k)
\label{eq:gcn_ours}
\end{equation}
where $K=3$ denotes the scale size of the spatial dimension, $W_k$ is the weight of the convolution operation, $X^l$ is the feature at the $l$-th layer, and $G_k$ is our proposed instance-dependent adjacency matrix generated from $X^l$. 

As shown in Figure~\ref{fig:pipeline}(d), STF firstly uses a three-layer convolutional module $\beta(\cdot)$ to extract the spatio-temporal embedding $\beta(X^l)$ of dimension $T\times N$ on input $X^l$. Then we apply a CNN with two convolutional layers $\alpha(\cdot)$ to convert the embedding to $G_k = \alpha(\beta(X^l)) $ of dimension $N\times N$. Now the embedding $\beta(X^l)$ encodes a data-dependent graph that learns a unique graph for each sample via $\alpha(\cdot)$, but it does not guarantee that $\beta(X^l)$ can represent the high-order topology and importance in the spatio-temporal domain. To address this issue, we propose to use backward gradient to model spatio-temporal focus to supervise the learning of the STF modules.

To extract the spatio-temporal focus to guide the learning, we apply the Grad-CAM \cite{selvaraju2017grad} to the last layer (which contains more abstract features and has a larger receptive field than the previous layers) and extract the spatio-temporal focus $Q$ as follows:
\begin{equation}
    Q = ReLU\left(\sum_c\left(\frac{1}{Z}\sum_{t,v}\frac{\partial \hat{y}}{\partial X^l_{ctv}}\right)X^l_{ctv}\right),
\label{eq:attention}
\end{equation}
where $\hat{y}$ is the probability of the predicted class $y$.
$c$, $t$, and $v$ denote the channel, temporal, and spatial dimensions of the intermediate feature map $X^l_{ctv}$, respectively, and $Z$ is the normalization factor in the spatio-temporal dimension. $Q$ is normalized to $[0,1]$ by the min-max normalization. Then we learn the spatio-temporal embedding network $\beta$ using $Q$ in Eq.~\ref{eq:attention} as supervision via $L_{G_k} = \|Q - \beta{(X^l)}\|_2$, which ensures that the spatio-temporal embedding encodes the instance-dependent and high-order spatio-temporal dependencies. As we use the spatio-temporal focus $Q$ as the guidance, the accuracy of $Q$ becomes critical. We propose the following three objectives to further regularize $Q$.
%global average pooling factor
%Then we learn the spatio-temporal embedding network $\beta(X^l)$

\subsection{STF Exploration}
\label{method:l_ae}

% Firstly, STF exploration is introduced to ensure the classifier's attention completely covers all the important spatio-temporal features to make prediction with a global view. 
Firstly, the STF exploration loss is introduced to ensure the complete and global view of the spatio-temporal focus $Q$. 
Typical classification losses, such as the cross-entropy loss used in \cite{krizhevsky2012imagenet}, do not enforce the classifier to infer based on all critical skeleton joints in all important frames. Thus there is no constraint that the classifier will look over the entire critical parts in the spatio-temporal space. The classifier which only pays attention to partial critical parts can predict incorrectly when such information is occluded or noisy~\cite{Li_PAMI19}.

Therefore, we propose the STF exploration loss $L_{e}$ to enforce the classifier's spatio-temporal focus to cover all the critical joints and all the critical frames.
% make inference based on all critical joints across all the critical frames that activate the predicted class. 
Specifically, suppose that the input sequence is classified to class $y$. An ideal classifier's focus should cover the entire critical parts across the input sequence. If we mask the corresponding parts of the spatio-temporal input sequence, the predicted probability of class $y$ should be as low as possible. We formulate this process as:
\begin{equation}
    L_{e} = g^y(\mathcal{X} - Q\odot \mathcal X),
\label{eq:l_ae}
\end{equation}
%where $g^y(\cdot)$ is the network $g$'s prediction score at the original prediction class $y$ in Eq.~\ref{eq:attention}
where $g^y(\cdot)$ extracts the prediction score of the spatio-temporal focus masked input $(\mathcal{X} - Q\odot \mathcal X)$ at the original prediction class $y$ in Eq.~\ref{eq:attention}, $\odot$ is the element-wise multiplication, and $\mathcal X$ is the input skeleton sequence. $L_{e}$ uses an exclusion strategy by eliminating $Q$'s corresponding spatio-temporal input to guide the network to focus on the critical parts across the input sequence. We observed that the focus $Q$ expands to include more critical joints in the spatio-temporal space than $Q$ without using $L_{e}$.
% Figure~\ref{fig:attention}(a)

% \begin{figure*}
% \centering{
% \includegraphics[width=.97\linewidth]{img/attention.png}}
% \vspace{-.5em}
% \caption{The spatio-temporal attention volumes before vs. after adding (a) $L_{e}$, (b) $L_{d}$, and (c) $L_{c}$. The x/y axis of each attention volume represents the temporal axis/joint ID, as shown in the bottom left figure. Unless otherwise specified, all attention volumes are the top-1 attention volumes from the GCN-6 module in Figure~\ref{fig:pipeline}. The \green{green}/\red{red} boundary of the top attention volumes of each example represents the correct/incorrect prediction from the classifier. We show more examples in the supplementary materials. Annotation: GT: ground truth.
% }
% \label{fig:attention}
% \vspace{-.65em}
% \end{figure*}

\subsection{STF Divergence}

%\kc{Avoid motivating from $L_{e}$ since we don't train with $L_{d}$ and $L_{e}$ jointly.}

%However, the $L_{e}$ introduced in Sec. \ref{method:l_ae} enforces the classifier to cover the critical parts as completely as possible, which can potentially make the attention volumes to include the entire input sequences, which is an obstacle for the classifier to separate the attention volumes associated with different classes.

% \kc{Rewrite this paragraph. The motivation of $L_{d}$ should be: 1. Preventing $L_{e}$ from making the attention maps blindly cover the entire sequence. 2. Inspired by [my ICCV19 paper], given an input sequence, the overlap of the attention maps associated with different classes causes confusion of the classifier. (Using citations to justify your motivation is stronger than using just your intuition (because the reviewers may disagree with your intuition).)}

%separate the attention maps of different classes
%to ensure that the difference of the classifier's attention to different classes reflects the difference of them.
Secondly, we propose the STF divergence loss to encourage the classifier to focus on different parts when predicting different classes.
%Secondly, we propose the STF divergence loss to ensure that the classifier's attention volumes of different classes are different.
The skeleton sequences of different classes usually involve either different joints or different temporal movements, or both. For example, the reading and walking classes involve different joints (the upper body and lower body parts). In contrast, the walking and running classes involve similar parts (legs and arms) but have different temporal patterns (legs/arms' movement is different). These differences can be represented by the differences of their spatio-temporal focus. 
Inspired by the finding that overlapped attention of different classes causes visual confusion~\cite{wang2019sharpen}, we propose the STF divergence loss $L_{d}$ to reduce the overlap of the focus of different classes. This loss discourages the network from covering all the parts across the sequence and focuses more on different classes' discriminative features. Although the concept of this loss can be applied to all the confusing classes, to simplify the optimization, we mainly focus on separating the most confusing classes (\eg, touching neck vs. touching head).
%walking and running
Specifically, given an input sequence, we select the top two predictions as the most confusing classes and enforce the top two classes' focus to overlap as little as possible.

Similar to $L_{e}$, we define $L_{d}$ with the spatio-temporal focus $Q$ via Eq. \ref{eq:attention} as: 
\begin{equation}
L_{d} = - ||Q^{y_i} - Q^{y_j}||_2,
\label{eq:l_ad}
\end{equation}
where $Q^{y_i}$ and $Q^{y_j}$ are the spatio-temporal focus of the top two prediction classes $y_i$ and $y_j$, respectively. 
% Figure~\ref{fig:attention}(b) 
We observed that by introducing $L_{d}$, the top two predictions' focus overlaps less. 
% The top example in Figure~\ref{fig:attention}(b) shows that by reducing the attention overlap of the most confusing classes, the misclassification from the classifier before using $L_{d}$ is successfully corrected.

\subsection{STF Coherence}

%\kc{Parts of the first paragraph are better to appear in Sec.~\ref{introduction} or Sec.~\ref{related_works}.}
%1. Stacked GCNs
%2. Later module has larger receptive field, thus contains global view 
%3. Operate on same sequence, thus should have similar attention 
Thirdly, we propose the STF coherence loss to utilize the high-level GCN module's focus to assist the low-level GCN module's learning.
In skeleton-based action recognition, stacked GCNs are commonly used network structure \cite{si2018skeleton,liu2020disentangling,Peng2020LearningGC,Shi2019TwoStreamAG}. The modules closer to the final output (high-level modules) capture more abstract information with a larger receptive field. The modules closer to the input (low-level modules) have relatively smaller receptive fields. Thus, the high-level modules' focus has a more global view of the input skeleton sequence than that from the low-level modules. We can use the focus of a more global view from the high-level module to guide the learning of the low-level modules. Moreover, the stacked GCNs operate on the same input sequence; thus, if there is an optimal focus for these networks to pay attention to, it will be more likely to be the focus of the high-level module. Therefore, we propose the STF coherence loss $L_{c}$, which enforces the stacked GCNs to have coherent focus across the network:
\begin{equation}
    L_{c} = ||Q_i - Q_j||_2,
\label{eq:l_ac}
\end{equation}
where $Q_i$ and $Q_j$ are the spatio-temporal focus from the STF modules $i$ and $j$, respectively. Ideally, the concept of $L_{c}$ can be applied to all the GCN modules in addition to the GCN modules $i$ and $j$. For the convenience and ease of optimization, we choose the last two GCN modules (STF-GCN-5 and STF-GCN-6 in Figure~\ref{fig:pipeline}) as the GCN modules $i$ and $j$ given that they capture more abstract information of the input sequence.

We qualitatively verify the efficacy of $L_{c}$
% in Figure \ref{fig:attention}(c), 
where we noticed that with $L_{c}$, the last two GCN modules' focus is more coherent than that without $L_{c}$, 
% Figure \ref{fig:attention} (c)
%In which, we also shows
and that enforcing the coherence of the focus from different GCN modules can correct the misclassification from the classifier which does not use $L_{c}$.

% \begin{table*}
% \small
% %\footnotesize
% \centering
% %\begin{tabular}{@{}c@{\hspace{.8em}}c@{\hspace{.8em}}cc@{\hspace{.8em}}c@{\hspace{.8em}}c@{}}
% \begin{tabular}{@{}cccccc@{}}
% \toprule
% %dataset notation &\multicolumn{2}{c}{$D_{n60}$} &\multicolumn{2}{c}{$D_{n120}$} &$D_k$\\
% property $\backslash$ dataset &\multicolumn{2}{c}{NTU RGB+D 60} &\multicolumn{2}{c}{NTU RGB+D 120} &Kinetics Skeleton 400\\
% \midrule
% settings &x-sub &x-view &x-set &x-sub &-\\
% \# class &60 &60 &120 &120 &400\\
% %\# sequence &56880 & &114480 & &306245\\
% \# training/testing sequences &40091/16487 &37646/18932 &54468/59477 &63026/50919 &240436/19796\\
% %\# training / testing sequences &40091 / 16487 &37646 / 18932 &54468 / 59477 &63026 / 50919 &240436 / 19796\\
% %\# total subjects &40 & &106 & &\\
% %\# training / testing subjects &20 / 20 &- & - &53 / 53 & -\\
% %\# training / testing views & - & 2 / 1 &- & - & -\\
% %\# training / testing setups &- &- & 16 / 16 & - & -\\
% \# training/testing subjects/views/setups &20/20 (subjects) &2/1 (views) &16/16 (setups) &53/53 (subjects) & -\\
% % \#Views      & 80           & 155          &    -      &  \\
% %\# people per sequence &$\le$2 &$\le$2 &$\le$2 &$\le$2 &$\le$2\\
% \# keypoints &25 &25 &25 &25 &18\\
% spatial coordinate &3D &3D &3D &3D &2D\\
% \bottomrule
% \end{tabular}
% \vspace{.4em}
% \caption{The statistics of the datasets we use. All the datasets have $\le$2 people per sequence.}
% \label{table:dataset}
% \vspace{-.7em}
% \end{table*}

\subsection{Overall STF Loss}
\label{method:overall_STF_loss}

We use our proposed STF modules and objectives together with the cross-entropy loss $L_{ce}$ for skeleton-based action recognition. Specifically, the overall loss is:
\begin{equation}
    L = L_{ce} + \lambda_{e} L_{e} + \lambda_{d} L_{d} + \lambda_{c} L_{c} + \lambda_{G_k}L_{G_k},
\end{equation}
where $\lambda_{e}, \lambda_{d}, \lambda_{c}, \lambda_{G_k}$ are the weights of the losses, so that each loss term has comparable absolute range. We separate $L_{e}$ from $L_{d}$, $L_{c}$, and $L_{G_k}$ during training, and ensemble them during testing because we find that optimizing $L_{e}$ and $L_{d}$ together makes the training process unstable. We hypothesize that it is because the goals of $L_{e}$ and $L_{d}$ can be conflicting implicitly -- $L_{e}$ tends to expand the focus to include all critical joints, but $L_{d}$ typically shrinks the focus to reduce the overlap of the focus corresponding to confusing classes. %More training and testing details are in the supplementary material.
% More training and testing details are in the experiment section and supplementary material.
% \kc{The readers may not understand what we mean by ``separate," especially when the following sentences describe the conflict of $L_{e}$ and $L_{d}$. It can be confusing for the readers how exactly we train STF. It may be necessary to describe the ensemble here or in the supplementary material.}, 

%\kc{It's strange that $L_{ce}$ appears in Eq. 6 twice without mentioning the ensemble. In the ``Joint-bone stream fusion" section, you still mention STF$_e$ and STF$_{dcg}$. Please be consistent.}

%To mitigate the conflict between $L_{e}$ and $L_{d}$, we train them separately, and ensemble their results as our final method STF (see the ``Joint-bone stream fusion" section).

% the overall loss $L_E$ for STF$_e$ is:
% \begin{equation}
% L_E = L_{ce} + \lambda_{e} L_{e},
% \end{equation}
% and the overall loss $L_{DCG}$ for STF$_{dcg}$ is:
% \begin{equation}
%     L_{DCG} = L_{ce} + \lambda_{d} L_{d} + \lambda_{c} L_{c} + \lambda_{G_k}L_{G_k},
% \end{equation}

%To train the network with the proposed objectives jointly with the cross-entropy loss $L_{cls}$ for action recognition, we form the final loss $L$ of our method GAAMS as follows:
%\begin{equation}
%    L = L_{cls} + \lambda_{ad} L_{d} + \lambda_{ac} L_{c} + \lambda_{G_k}L_{G_k}
%\end{equation}
%where $\lambda_{ae}, \lambda_{ad}, \lambda_{ac}, \lambda_{G_k}$ are the weights that balance the losses.

% \kc{Why don't you change the notations $\alpha$ in the equation?}

%\section{Experimental setup}
\section{Experiment}
\label{experiments}

\begin{table*}[t]
\centering
%\small
\footnotesize
%\scriptsize
\resizebox{2.1\columnwidth}{!}{
%\begin{tabular}{@{}rc@{\hspace{.7em}}c@{\hspace{.7em}}c@{\hspace{2em}}c@{\hspace{.7em}}c@{\hspace{.7em}}c@{\hspace{2em}}c@{\hspace{.7em}}c@{\hspace{.7em}}c@{\hspace{2em}}c@{\hspace{.7em}}c@{\hspace{.7em}}c@{}}
\begin{tabular}{@{}r@{\hspace{.35em}}c@{\hspace{.35em}}c@{\hspace{.35em}}c@{\hspace{.55em}}c@{\hspace{.35em}}c@{\hspace{.35em}}c@{\hspace{.55em}}c@{\hspace{.35em}}c@{\hspace{.35em}}c@{\hspace{.55em}}c@{\hspace{.35em}}c@{\hspace{.35em}}c@{\hspace{.55em}}c@{\hspace{.35em}}c@{\hspace{.35em}}c@{}}
\toprule
dataset &\multicolumn{6}{c}{NTU RGB+D 60} &\multicolumn{6}{c}{NTU RGB+D 120} &\multicolumn{3}{c}{Kinetics-400}\\
\midrule
setting &\multicolumn{3}{c}{\hspace{-1em}x-sub} &\multicolumn{3}{c}{\hspace{-1.2em}x-view} &\multicolumn{3}{c}{\hspace{-1.4em}x-sub} &\multicolumn{3}{c}{x-set}\\
method $\backslash$ input modality &$J$ &$B$ &$J$+$B$ &$J$ &$B$ &$J$+$B$ &$J$ &$B$ &$J$+$B$ &$J$ &$B$ &$J$+$B$ &$J$ &$B$ &$J$+$B$\\
\midrule
% ST-GCN~\cite{Yan2018SpatialTG} $_{\text{AAAI'18}}$ &81.50 &- &- &88.30 &- &- &- &- &- &- &- &- &- &- &-\\
SR-TSL~\cite{si2018skeleton} $_{\text{ECCV'18}}$ &84.80 &- &- &92.40 &- &- &- &- &- &- &- &- &- &- &-\\
% AS-GCN~\cite{li2019actional} $_{\text{CVPR'19}}$ &86.80 &- &- &94.20 &- &- &- &- &- &- &- &- &- &- &-\\
% AGC-LSTM~\cite{Si_2019_CVPR} $_{\text{CVPR'19}}$ &87.50 &- &89.20\dag &93.50 &- &95.00\dag &- &- &- &- &- &- &- &- &-\\
% 2s-NLGCN~\cite{Shi2018NonLocalGC} $_{\text{arXiv'18}}$ & & &88.50 &93.70 &93.20 &95.10 &- &- &- &- &- &- &- &- &-\\
2s-AGCN~\cite{Shi2019TwoStreamAG} $_{\text{CVPR'19}}$ &- &- &88.50 &93.70 &93.20 &95.10 &- &- &- &- &- &- &35.10 &33.30 &36.10\\
% ST-GST~\cite{pan2021spatio} $_{\text{ICLR'21}}$ &73.10 &- &- &- &- &- &- &- &- &- &- &- &- &- &-\\
TS-SAN~\cite{cho2020self} $_{\text{WACV'20}}$ &87.20 &- &- &92.70 &- &- &- &- &- &- &- &- &35.10 &- &-\\
GCN-NAS~\cite{Peng2020LearningGC} $_{\text{AAAI'20}}$ &- &- &89.40 &94.60 &94.70 &95.70 &- &- &- &- &- &- &35.50 &34.90 &37.10\\
MS-TGN~\cite{li2020multi} $_{\text{ICCSIT'20}}$ &86.60 &87.50 &89.50 &94.10 &93.90 &95.90 &- &- &- &- &- &- &35.20 &33.30 &37.30\\
MS-AAGCN~\cite{Shi2019SkeletonBasedAR} $_{\text{TIP'20}}$ &88.00 &88.40 &89.40 &95.10 &94.70 &96.00 &- &- &- &- &- &- &36.00 &34.70 &37.40\\
3s RA-GCN~\cite{song2020richly} $_{\text{TCSVT'20}}$ &- &- &87.30$\star$ &- &- &93.60$\star$ &- &- &81.10$\star$ &- &- &82.70$\star$ &- &- &-\\
DC-GCN+ADG~\cite{Cheng_ECCV20} $_{\text{ECCV'20}}$ &- &- &90.80\ddag &- &- &96.60\ddag &- &- &86.50\ddag &- &- &88.10\ddag &- &- &-\\
DSTA-Net~\cite{shi2020decoupled} $_{\text{ACCV'20}}$ &- &- &91.50$\diamond$ &- &- &96.40$\diamond$ &- &- &86.60$\diamond$ &- &- &89.00$\diamond$ &- &- &-\\
%$M_{11}$ &DSTA~\cite{shi2020decoupled} $_{\text{arXiv'20}}$ &- &- &91.50 &- &- &96.40 &- &- &86.60 &- &- &89.00\\
STIGCN~\cite{huang2020spatio} $_{\text{ACMMM'20}}$ & 90.10& - &-&96.10&-&-&- &- &- &- &- &- &37.90 &- &-\\
PA-ResGCN-B19~\cite{song2020stronger} $_{\text{ACMMM'20}}$ &- &- &90.90\dag &- &- &96.00\dag &- &- &87.30\dag &- &- &88.30\dag &- &- &-\\
% ST-TR~\cite{plizzari2020spatial} $_{\text{ICPRW'20}}$ &88.70 &- &89.90 &95.60 &- &96.10 &81.90 &- &- &84.10 &- &- &- &- &-\\
%ST-BLN~\cite{Heidari2020stbln} $_{\text{arXiv'20}}$ &85.71 &- &87.80 &93.80 &- &95.10 &- &- &- &- &- &- &- &- &-\\
PST-GCN~\cite{Heidari2020pstgcn} $_{\text{arXiv'20}}$ &87.90 &- &88.68 &94.33 &- &95.10 &- &- &- &- &- &- &34.71 &- &35.53\\
Dynamic-GCN~\cite{ye2020dynamic} $_{\text{ACMMM'20}}$ &- &- &91.50 &- &- &96.00 &- &- &87.30 &- &- &88.60 &- &- &37.90\\
% Yang \etal~\cite{yang2020improving} $_{\text{arXiv'20}}$ &86.70 &- &- &93.30 &- &- &83.80 &- &- &85.70 &- &- &32.10 &- &-\\
MS TE-GCN~\cite{li2020temporal} $_{\text{arXiv'20}}$ &87.40 &88.50 &90.80\ddag &93.40 &93.30 &96.20\ddag &- &- &84.40\ddag &- &- &85.90\ddag &- &- &-\\
%KShapeNet~\cite{friji2020kshapenet} $_{\text{arXiv'20}}$ &97.20 &- &- &96.20 &- &- &64.00 &- &- &66.10 &- &-\\
% HCSF~\cite{zeng2021learning}$_{\text{ICCV'21}}$ &- &- &91.6& &- &-	96.7	&- &- &87.5	&- &- &89.2 & - &- &- \\
% Alsawadi's~\cite{Alsawadi_arXiv21} $_{\text{arXiv'21}}$  &- &- &82.60	 &- &- &90.50	 &- &- &-	 &- &- &-  &- &- &31.70  \\
% STAR~\cite{Shi_arXiv21} $_{\text{arXiv'21}}$  &- &- &83.40	 &- &- &89.00	 &- &- &78.30	 &- &- &80.20  &- &- &-  \\
UNIK~\cite{Yang_arXiv21} $_{\text{arXiv'21}}$  &- &- &86.80	 &- &- &94.40	 &- &- &80.80	 &- &- &86.50  &- &- &-  \\
AdaSGN~\cite{shi2021adasgn} $_{\text{arXiv'21}}$  &- &- &89.10	 &- &- &94.70	 &- &- &85.90	 &- &- &86.80  &- &- &-  \\
Yang's~\cite{Yang_ICCV21} $_{\text{ICCV'21}}$ &88.00 &- &- 	 &94.90 &- &- 	 &- &- &-	 &- &- &-  &- &- &-  \\
\gray{MS-G3D (paper)~\cite{liu2020disentangling} $_{\text{CVPR'20}}$} &\gray{89.40} &\gray{90.10} &\gray{91.50} &\gray{95.00} &\gray{95.30} &\gray{96.20} &- &- &\gray{86.90} &- &- &\gray{88.40} &\gray{35.80} &\gray{35.44} &\gray{38.00}\\
MS-G3D (code)*~\cite{liu2020disentangling} $_{\text{CVPR'20}}$ &88.77 &89.59 &90.67 &94.88 &94.86 &95.82 &82.35 &84.86 &86.42 &84.14 &86.79 &87.98 &35.74 &34.77 &37.23\\
\midrule
% STF$_e$ &89.17 &89.70 &90.94 &94.93 &94.91 &95.99 &82.91 &85.20 &86.59 &84.40 &87.19 &88.16 &36.77 &36.34 &38.41\\
% STF$_{dcg}$ &89.29 &89.84 &91.02 &94.97 &95.11 &96.13 &83.32 &84.93 &86.55 &84.72 &87.15 &88.22 &37.25 &36.72 &38.40\\
%\midrule
% STF (ours) &\textbf{90.74} &\textbf{91.09} &\textbf{91.88} &\textbf{96.32} &\textbf{95.87} &\textbf{96.61} &\textbf{84.67} &\textbf{85.88} &\textbf{87.64} &\textbf{86.02} &\textbf{88.71} &\textbf{89.04} &\textbf{38.20} &\textbf{37.56} &\textbf{39.87}\\

 STF (ours) &\textbf{91.34} &\textbf{91.09} &\textbf{92.47} &\textbf{96.46} &\textbf{96.51} &\textbf{96.86} &\textbf{85.06} &\textbf{86.80} &\textbf{88.85} &\textbf{86.40} &\textbf{88.86} &\textbf{89.92} &\textbf{38.20} &\textbf{37.56} &\textbf{39.87}\\

\bottomrule
\end{tabular}
}
\vspace{-0.5em}
\caption{Performance comparison of skeleton-based action recognition in top-1 accuracy (\%). *: For MS-G3D~\cite{liu2020disentangling}, the publicized code~\cite{MSG3D_code} we use as our baseline has lower accuracy than what was reported in their paper. %Since the  backbone and code base of STF are built on top of the publicized code~\cite{MSG3D_code} of MS-G3D, we report the result obtained using the publicized code~\cite{MSG3D_code} of MS-G3D for fair comparison. 
Annotations: $J$: joint; $B$: bone; ``-": results not provided in the reference. The methods requiring more inputs than $J$+$B$: ``$\star$":~\cite{song2020richly} uses 3 streams; ``\ddag":~\cite{Cheng_ECCV20,li2020temporal} use $J$+$B$+$J$ motion+$B$ motion; ``$\diamond$":~\cite{shi2020decoupled} uses spatial-temporal, spatial, slow-temporal, and fast-temporal streams; ``\dag":~\cite{song2020stronger} uses $J$+$B$+velocity. %\kc{Please replace HCSF with other methods published in 2021.} %``$\triangleleft$":~\cite{li2020temporal} uses $J$+$B$+$J$ motion+$B$ motion.
}
\label{table:exp_result}
\vspace{-.5em}
\end{table*}

\subsection{Datasets}

% //NTU RGB-D 60 \cite{NTU60}, NTU RGB-D 120 \cite{NTU120}, Kinetics Skeleton 400 \cite{Kinetics}, OpenPose \cite{OpenPose}

%To evaluate the effectiveness of the proposed method, we perform experiment on the following three datasets:

We conduct experiments on three benchmark datasets, namely, the NTU RGB+D 60~\cite{NTU60}, NTU RGB+D 120~\cite{NTU120}, and Kinetics Skeleton 400~\cite{Kinetics} datasets (denoted as NTU-60, NTU-120, and Kinetics-400, respectively). %The statistics of these datasets can be found in the supplementary material.
% We summarize the statistics of these datasets in Table~\ref{table:dataset}.
%Some detailed statistics of these datasets are summarized in Table~\ref{table:dataset}.

{\bf NTU RGB+D 60}~\cite{NTU60} is a large-scale skeleton-based action recognition dataset with over 60 action classes of 40 subjects for indoor scenarios. Each sequence contains one or two persons' 3D skeletons captured by three Kinect v.2 cameras in three views (termed as views 1, 2, and 3). The recommended two settings are (1) cross-Subject (x-sub), where the dataset is equally split as training and testing sets of 20 subjects each; and (2) cross-View (x-view), where all samples from view 1 are used for testing and the samples from views 2/3 are used for training.
%In addition, we experiment on the sub-sampled dataset (\ie, train on 10/20/25\% randomly sampled training set and test on the entire testing set) under the x-sub setting to simulate the potential data scarcity issue in practice. 
%The recommended reporting classification accuracy has two settings

{\bf NTU RGB+D 120}~\cite{NTU120} extends the NTU-60 dataset with 60 extra action classes, resulting in a total of 120 action classes, for 106 subjects. The recommended two settings are (1) cross-Subject (x-sub), the 106 subjects are split into 53/53 subjects for training/testing; and (2) cross-Setup (x-set), where 16/16 setups are used for training/testing.
%16 setups are used for training, and the other 16 setups are reserved for testing.

{\bf Kinetics Skeleton 400} \cite{Kinetics} is a skeleton-based action recognition dataset converted from the Kinetics 400 video dataset \cite{Kinetics} using the OpenPose \cite{OpenPose} toolbox in 2D keypoints modality.

\subsection{Implementation Details and Protocols}
\label{exp:protocol}
% \lk{data preprocessing: 4D}

%The proposed model are tuned from a baseline model
%tune our proposed objectives on the baseline model
%Inputs are preprocessed with normalization and translation following \cite{si2018skeleton}.
{\bf Data preparation}. We preprocess the input skeleton sequences by subtracting each joint position by the center joint position and normalizing the results by the body height, in the same fashion as \cite{si2018skeleton}. No other data processing or augmentation is used for fair comparisons. All skeleton sequences are padded to $T$=300 frames by repeating the sequences. To keep the semantic information of the skeleton sequence in Eq.~\ref{eq:l_ae}, as done in \cite{si2018skeleton}, we add a visibility channel alongside the $X$-$Y$-$Z$ location channel to indicate the joint's visibility, where 1/0 means visible/invisible. The initial visibility channel is set to 1. The visibility channel is applied to the NTU-60 and NTU-120 datasets. For the Kinetics-400 dataset, the visibility is embedded with the confidence score channel of joints. Unless otherwise specified, we use the default settings for other parameters. %More implementation details are in the supplementary materials.

{\bf Experimental protocol}. 
Following the experimental protocol of Shi \etal~\cite{Shi2019TwoStreamAG}, we report the classification accuracies under three different input modalities: (i) $J$: the joints only, where the joint stream is the absolute skeleton joints location sequence, (ii) $B$: the bones only, where the bone stream is the second-order information (the lengths and directions of bones) of the skeleton data, and (iii) $J$+$B$: the 2-stream input with both joints and bones.
%~\cite{Yan2018SpatialTG,si2018skeleton,li2019actional,Si_2019_CVPR,Shi2019TwoStreamAG,Peng2020LearningGC,cho2020self,Shi2019SkeletonBasedAR,Cheng_ECCV20,Heidari2020stbln,Heidari2020pstgcn,liu2020disentangling}.  

{\bf Training scheme}. We implement STF using MS-G3D~\cite{liu2020disentangling} as the backbone. The MS-G3D baseline model is trained using SGD with momentum 0.9, batch size 32, initial learning rate 0.05, and weight decay 0.0005, and the base learning rate is adjusted accordingly for different settings. For NTU-60, NTU-120, and Kinetics-400, the learning rate is decayed at [20, 35, 45], [20, 35, 50], [25, 40, 55] epochs, respectively. After that, we pre-train the STF model from the MS-G3D baseline model, with lower initial learning rates $\{10^{-3}, 5\times10^{-4}, 10^{-4}\}$. We empirically set $\lambda_{e}$/$\lambda_{d}$/$\lambda_{c}$/$\lambda_{G_k}$ as 0.01/0.1/0.1/0.01, respectively, such that all loss terms have comparable absolute ranges. %the absolute ranges of all loss terms are comparable.
%we use the pre-trained STF model from the MS-G3D baseline model

%Following \cite{si2018skeleton}, we preprocess the inputs with normalization and translation.

{\bf Joint-bone stream fusion}. Moreover, we follow Shi \etal~\cite{Shi2019TwoStreamAG} to combine the results of joint and bone streams by averaging the prediction probability for each sequence, and report them in Table~\ref{table:exp_result}. Specifically, 
% the results of joint ($J$) and bone ($B$) streams for STF in Table~\ref{table:exp_result} are fused from $J$ and $B$ streams, respectively. We
we merge the $J$ and $B$ streams by averaging the prediction probability from both streams to get the $J$+$B$ results of STF. % \kc{STF$_e$ and STF$_{dcg}$ appear without explanation.}
%the results of joint ($J$) and bone ($B$) streams for ensemble (STF$_e$, STF$_{dcg}$) of Table~\ref{table:exp_result} are fused from $J$ streams of STF$_e$ and STF$_{dcg}$

\subsection{Experimental Results}
\label{result}

% \kc{You may want to briefly explain the meaning of joint only, bone only, and joint + bone. Please explain our ensemble methods ($M_{16}$, $M_{17}$).}

{\bf Baselines}. We summarize the experimental results in Table~\ref{table:exp_result}, where we compare STF with the SOTA skeleton-based action recognition methods. %~\cite{pan2021spatio,cho2020self,Peng2020LearningGC,li2020multi,Shi2019SkeletonBasedAR,song2020richly,Cheng_ECCV20,shi2020decoupled,huang2020spatio,song2020stronger,plizzari2020spatial,Heidari2020stbln,Heidari2020pstgcn,yang2020improving,li2020temporal,liu2020disentangling}. 
Of all the methods in Table~\ref{table:exp_result}, MS-G3D~\cite{liu2020disentangling} is one of the most competitive methods, so we build STF on top of it.
%thus we choose it as our baseline model.
% and DC-GCN+ADG~\cite{Cheng_ECCV20} are the most competitive ones. 
However, when we run the publicized code \cite{MSG3D_code} of MS-G3D, we found that its accuracy is lower than that reported in the paper~\cite{liu2020disentangling}. Since we implement STF based on the publicized code of MS-G3D, we use the accuracy obtained from running the publicized code of MS-G3D for fair comparisons. We also include the performance reported in the original paper of MS-G3D~\cite{liu2020disentangling} in Table~\ref{table:exp_result}, where we mark the best performance in \textbf{bold}.

{\bf Comparison with SOTAs}. Table~\ref{table:exp_result} shows that STF outperforms the listed baselines in all settings.
% , and that STF outperforms STF$_e$, STF$_{dcg}$, and all the listed methods in all 15 settings. 
We find that although 3s RA-GCN~\cite{song2020richly}, DC-GCN+ADG~\cite{Cheng_ECCV20}, DSTA-Net~\cite{shi2020decoupled}, PA-ResGCN-B19~\cite{song2020stronger}, and MS TE-GCN~\cite{li2020temporal} have the unfair advantage of using strictly more information than $J$+$B$ as input, STF still outperforms them in most settings. In addition, we compare the accuracy gains of STF and MS-G3D over their baselines in Table~\ref{perf_improvement}. We find that the accuracy gain of STF is 3.8/1.9/2.3 times of that of MS-G3D on the challenging Kinetics-400 dataset, given that STF only uses 10\% more FLOPs than MS-G3D.
%We also find that although DC-GCN+ADG~\cite{Cheng_ECCV20} has the unfair advantage of using strictly more information as input (\ie, DC-GCN+ADG~\cite{Cheng_ECCV20} uses 2 more streams---motion and bone motion---in addition to $J$+$B$), STF$_e$ and STF$_{dcg}$ still outperform DC-GCN+ADG in 3 out of 4 settings. 

\begin{table}
\small
\centering
\resizebox{1\columnwidth}{!}{
\begin{tabular}{@{}c@{\hspace{.5em}}c@{\hspace{.5em}}c@{\hspace{.5em}}c@{}}
\toprule
%settings & NTU-60 x-view-$J$ & Kinetics-$J$ &  Kinetics-$B$\\
settings & Kinetics-$J$ &  Kinetics-$B$ & Kinetics-$J$+$B$\\
\midrule
\textbf{$\Delta($}MS-G3D, 2s-AGCN$)$ &0.64\% (1.0x) &	1.47\%  (1.0x) &	1.13\% (1.0x) \\
\textbf{$\Delta($}STF, MS-G3D$)$ &\textbf{2.46}\%  (3.8x) &\textbf{2.79}\% (1.9x) &	\textbf{2.64}\% (2.3x)  \\%&1.97 &2.5 &2.8

%\textbf{$\Delta($}MS-G3D, 2s-AGCN$)$ &1.18\% &0.64\%&1.47\% \\
%\textbf{$\Delta($}STF, MS-G3D$)$ &\textbf{1.44}\% &\textbf{2.46}\% &\textbf{2.79}\%\\%&1.97 &2.5 &2.8
\bottomrule
\end{tabular}
}
\vspace{-.5em}
\caption{Comparison of accuracy gain over the baseline.}
\label{perf_improvement}
\end{table}

\begin{comment}
{\bf Analysis on STF variants}. Besides, we observe that STF$_{dcg}$ outperforms STF$_e$ in 11 out of 15 settings. We hypothesize that it is because: (1) STF$_{dcg}$ has more loss terms than STF$_e$ that correspond to additional information for learning, and (2) $L_d$ used to train STF$_{dcg}$ separates different classes via their attention volumes, which is more aligned with the goal of our task of interest -- differentiating different classes. Moreover, we observe that the improvement of STF under individual $J$ and $B$ streams is higher than that of STF$_e$ and STF$_{dcg}$. We speculate that this may be due to the conflict of STF$_e$ and STF$_{dcg}$ mentioned in Sec.~\ref{method:overall_STF_loss}, which increases the diversity in the learned features. On the other hand, the $J$ and $B$ streams in STF have higher correlations, thus when they are combined in $J$+$B$, they lead to smaller improvement.
%Moreover, we observe that the improvement of ensemble(STF$_a$, STF$_b$) under individual $J$ and $B$ streams is higher than that of STF$_a$ and STF$_b$. We speculate that this may be due to the conflict in STF$_a$ and STF$_b$ mentioned in Sec. \ref{sec:gk}, which increases the diversity in the learned features. On the other hand, the $J$ and $B$ streams in ensemble(STF$_a$, STF$_b$) have higher correlations, thus when they are combined in $J$+$B$, they lead to smaller improvement.
%Moreover, we observe that the improvement of $J$+$B$ over individual $J$ and $B$ streams is higher for STF$_a$ and STF than the case of ensemble (STF, STF$_a$).

\end{comment}

\begin{figure}
% \centering
\includegraphics[width=.99 \linewidth]{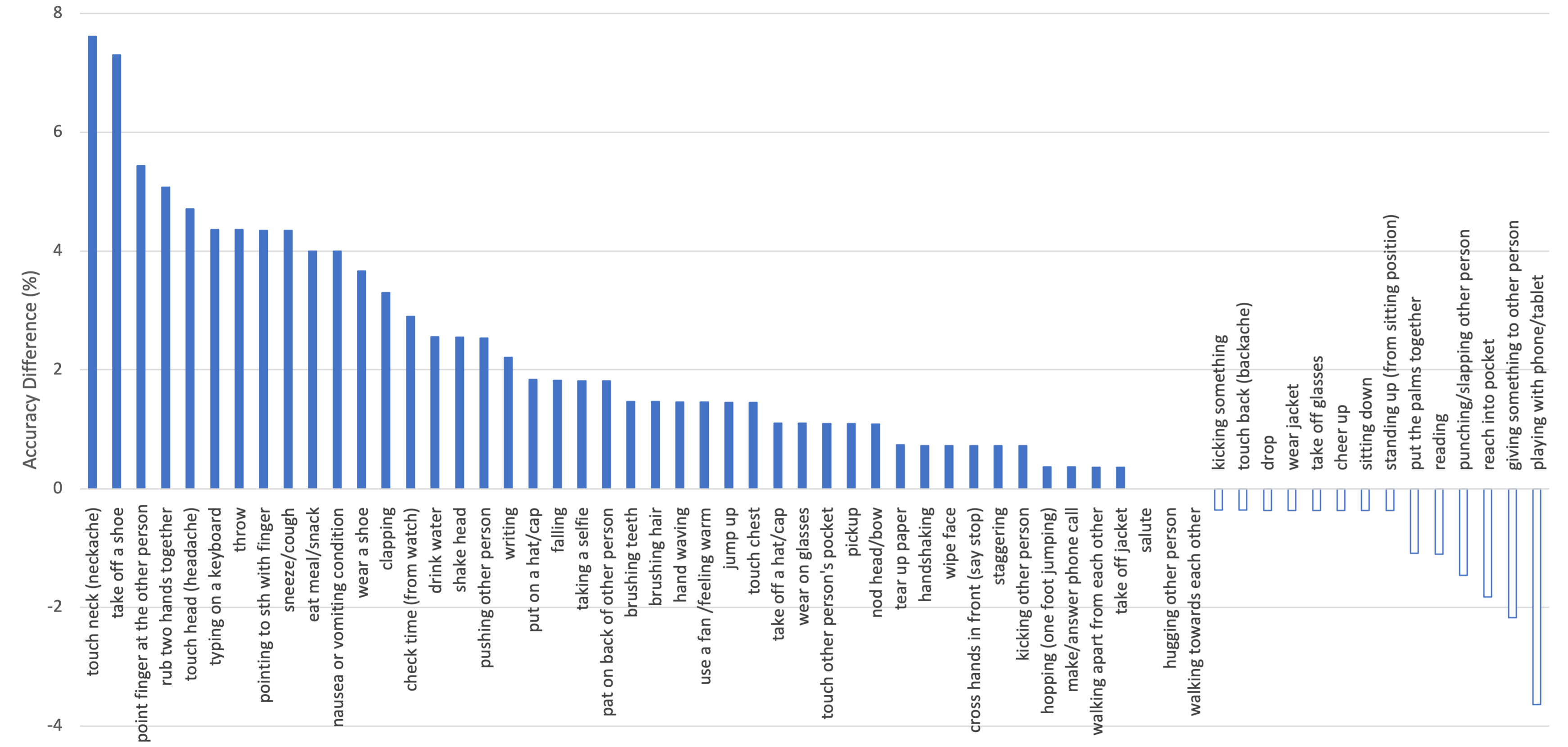}
\vspace{-.3em}
\caption{The class-wise accuracy difference (\%) between STF %(ensemble(STF$_e$, STF$_{dcg}$) 
in Table~\ref{table:exp_result}) and the baseline MS-G3D~\cite{liu2020disentangling} (MS-G3D (code) in Table~\ref{table:exp_result}) for the NTU-60 dataset under the x-sub setting with the joint input modality.
}
\label{fig:accuracy_diff}
\vspace{-.7em}
\end{figure}

{\bf Improvement on scarce data}. To simulate the real-world scenarios where only small amount of training data are available, we train on 10/20/25\% of randomly sampled training data from NTU-60 under the x-sub setting with the joint input modality, and evaluate the accuracy on the entire testing set of NTU-60. 
% Since MS-G3D~\cite{liu2020disentangling} is the most competitive single-stream baseline according to Table~\ref{table:exp_result}, we only compare with MS-G3D in
Table~\ref{table:exp_diff_data_amount} shows that STF consistently outperforms the baseline in all settings.
%Table~\ref{table:exp_diff_data_amount} shows that STF$_e$, STF$_{dcg}$, and STF consistently outperform the baseline in all settings.
% Table~\ref{table:exp_diff_data_amount} shows that STF outperforms STF$_a$ in all settings and that our ensemble method outperforms STF$_a$ and STF$_b$, which is consistent with Table~\ref{table:exp_result}. 
%We also verify this on Kinetics-400 in Table~\ref{table:exp_kinetics}, where we show that STF also outperforms MS-G3D, regardless of the input modality.
We also verify this on the randomly sampled 25\% Kinetics-400, the STF (joint 25.56, bone 26.30) outperforms MS-G3D (joint 24.69, bone 25.22), regardless of the input modality. 

{\bf Efficacy under data shift}.
%In Table~\ref{table:dataset_shift}, we also show that the features learned from STF outperform those learned from MS-G3D in the dataset shift setting, where we freeze the features learned from NTU-60, train a linear classifier on top for NTU-120, and evaluate on NTU-120. The results support that the features learned from our proposed STF$_e$, STF$_{dcg}$, and STF have better generalization ability than those learned from MS-G3D.
To test the efficacy of the features learned from our proposed STF across different datasets, we test STF model on the dataset different from the training set. Specifically, we freeze the model trained on the NTU-60 dataset under the x-sub setting with joint input modality, and replace the output fully connected layer with a new fully connected layer of 120 output nodes. The new fully connected layer is fine-tuned using the training set of the NTU-120 under the x-sub setting with joint input modality. After fine-tuning, we evaluate its performance on the NTU-120 testing set. Our experimental result shows that STF outperforms the baseline MS-G3D by 1.04\% (83.38\% vs. 82.34\%). This result supports that the features learned from STF has better generalization ability than MS-G3D.

\begin{table}[t]
\small
\centering
\resizebox{1\columnwidth}{!}{
\begin{tabular}{@{}r@{\hspace{.5em}}c@{\hspace{.7em}}c@{\hspace{.7em}}c@{\hspace{.7em}}c@{}}
\toprule
method $\backslash$ $p$ &10 &20 &25 &100\\
\midrule
%2s-AGCN~\cite{Shi2019TwoStreamAG} & & &79.18 / 95.35 &87.97 / 97.70\\
MS-G3D (code) &72.01 &79.11 &81.91 &88.77\\
%\midrule
% STF$_e$  &72.29 &79.18 &82.03 &89.17\\
% STF$_{dcg}$ &72.53 &80.61 &84.14 &89.29\\
%\midrule
STF & \textbf{72.73} ($\uparrow$0.72) & \textbf{80.77} ($\uparrow$1.66) & \textbf{84.27} ($\uparrow$2.36)&\textbf{91.34} ($\uparrow$2.57)\\
\bottomrule
\end{tabular}
}
\caption{The accuracy (\%) using joint input modality on the NTU-60 dataset under the x-sub setting. We use $p$\% of the randomly sampled training data from the NTU-60 dataset.}
\label{table:exp_diff_data_amount}
\vspace{-.5em}
\end{table}

% \begin{table}
% \small
% \centering
% \begin{tabular}{ccccc}
% \toprule
% method &MS-G3D &STF$_e$ &STF$_{dcg}$  &STF\\
% \midrule
% Accuracy &82.34 &82.61 & 82.76 & \textbf{83.38}\\
% \bottomrule
% \end{tabular}
% \vspace{-.5em}
% \caption{The accuracy (\%) of the experiment in the dataset shift setting (NTU-60$\rightarrow$NTU-120). \kc{Remove the table and just describe the result in the main text. If we have the cross-backbone results, please combine them here.} %The features learned from STF outperform those learned from MS-G3D~\cite{liu2020disentangling}, which supports that our proposed methods are beneficial in the dataset shift setting.
% }
% \label{table:dataset_shift}
% \vspace{-.5em}
% \end{table}

%\begin{table}
%\centering
%\begin{tabular}{ccc}
%\toprule
%method $\backslash$ input modality &joint &bone\\
%\midrule
%MS-G3D~\cite{liu2020disentangling} &24.69 &25.22\\
%STF &\textbf{25.56} &\textbf{26.30}\\
%\bottomrule
%\end{tabular}
%\vspace{.5em}
%\caption{The accuracy (\%) using joint and bone input modalities on the Kinetics-400 dataset.}
%\label{table:exp_kinetics}
%\vspace{-.5em}
%\end{table}

{\bf Improvement on confusing classes}. In addition, we also show a break-down evaluation of the class-wise accuracy difference (\%) for the joint input modality between STF %(ensemble(STF$_e$, STF$_{dcg}$) 
in Table~\ref{table:exp_result}) and the baseline MS-G3D~\cite{liu2020disentangling} (MS-G3D (code) in Table~\ref{table:exp_result}) for the x-sub setting on NTU-60. Figure \ref{fig:accuracy_diff} shows that STF outperforms MS-G3D for most classes. The highest performance gain occurs among the more challenging classes with more multiple joint correlations (\eg, touch neck $+$7.61\% vs. touch head $+$4.71\%, and take off a shoe $+7.30$\% vs. wear a shoe $+3.66$\%). This is likely due to the explicit consideration of the high-order dependencies in STF.

%especially on the confusing classes (\eg touch head $+$7.61\% vs. touch neck $+$4.71\%; wear shoe $+7.30$\% vs. take off shoe $+3.66\%$), which shows the effectiveness of the proposed attention guided losses.

\subsection{Ablation Study}
\label{ablation}

Our learning objectives are composed of several different loss terms, so it is important to know the contribution of these losses to accuracy.
%useful to know if all these terms are equally important to the final performance.
To this end, we perform an ablation study of the contribution of different loss terms using the joint input modality on the NTU-60 dataset under the x-sub setting in Table~\ref{ablation_exp}. We report the accuracy and $\Delta$, the accuracy improvement over MS-G3D~\cite{liu2020disentangling} by using different combinations of loss terms as the learning objectives. STF results in $\Delta=2.57$\% improvement with joint only modality on the NTU-60 x-sub setting.
%These results show that STF$_{dcg}$, which combines $L_{d}$, $L_{c}$ and $L_{G_k}$, leads to $\Delta=0.50$\%. STF$_e$ with $L_e$ gives $\Delta=0.38$\% performance gain. STF, the ensemble of STF$_e$ and STF$_{dcg}$, results in $\Delta=1.97$\% improvement with joint only modality on NTU-60 x-sub setting.
% and that dropping any term in the learning objective will not achieve the maximal performance improvement.

%These results show that $L_{c}$, which combines all three proposed loss terms, leads to the largest performance improvement, and dropping any term in the learning objective will not achieve the maximal performance improvement.

%Although we implement STF using MS-G3D~\cite{liu2020disentangling} as the backbone, our proposed $L_{e}$, $L_{d}$, $L_{c}$, and $L_{G_k}$ are not specifically tied to MS-G3D, and we believe that these losses have the potential to be applicable to other backbone architectures.

\begin{table}
\small
\centering
\begin{tabular}{@{}l@{}c@{\hspace{.65em}}c@{\hspace{.65em}}c@{\hspace{.65em}}c@{\hspace{.65em}}c@{\hspace{.65em}}c@{\hspace{0em}}c@{}}
\toprule
method &MS-G3D &$L_{c}$ &$L_{d}$ &$L_{G_k}$ &$L_{e}$ &accuracy &$\Delta$\\
\midrule
$M_1$&\checkmark & & & & &88.77 & ---\\ 
$M_2$&\checkmark & \checkmark & & & &89.33 &$\uparrow$ 0.56\\ 
$M_3$&\checkmark &\checkmark & \checkmark & & &89.92 &$\uparrow$ 1.15\\
$M_4$&\checkmark &\checkmark &\checkmark & \checkmark & &90.65 &$\uparrow$ 1.88\\ 
%$M_4$&\checkmark &\checkmark &\checkmark &\checkmark & &89.29 &$\uparrow$ 0.52\\
$M_5$: STF&\checkmark &\checkmark &\checkmark &\checkmark &\checkmark &\textbf{91.34} &$\uparrow$ 2.57\\
\bottomrule
\end{tabular}
\vspace{-.5em}
\caption{Ablation study of top-1 accuracy (\%) using joint only modality on the NTU-60 dataset under the x-sub setting. $\Delta$ shows the accuracy improvement over the baseline, MS-G3D~\cite{liu2020disentangling}. %\kc{Try to find whether we have ablation study on the Kinetics dataset.}
}
%STF is the ensemble of STF$_e$ and STF$_{dcg}$.
\label{ablation_exp}
\vspace{-.5em}
\end{table}

\section{Conclusion}
\label{conclusion}

%We proposed four spatio-temporal attention based objectives to build a novel spatio-temporal attention guided network.
%applies attention supervision on the embeddings to learn an attention guided adjacency matrix
%and attention supervision signal provide more spatio-temporal attention comparing with the unsupervised attention based adjacency matrix. Moreover, we use spatio-temporal attention to explicitly guide the learning process and improved the accuracy of the classifier.
%The experiment on NTU-60, NTU-120, and sampled Kinetics dataset shows the effectiveness of the proposed objectives.

We propose the To-a-T Spatio-Temporal Focus (STF) method for skeleton-based action recognition. First, we propose the STF modules to generate flexible adjacency matrices, and use the spatio-temporal focus to guide the learning of STF modules, such that the adjacency matrices capture the high-order dependency and spatio-temporal importance.
To capture critical spatio-temporal features, we propose the STF exploration, STF divergence, and STF coherence losses to encourage the spatio-temporal focus which supports the classifier's prediction to include all critical spatio-temporal features, to distinguish different classes on spatio-temporal focus, and to make the spatio-temporal focus across the stacked GCNs consistent, respectively. STF outperforms the SOTA methods on the NTU RGB+D 60, NTU RGB+D 120, and Kinetics Skeleton 400 datasets. Our proposed objectives in STF are not tied to any specific network architecture and have the potential to be applied to other network architectures, which is our future work. We also plan to explore multi-stream input to train all our proposed objectives jointly in the future.
%We also plan to explore other methods to train all our proposed objectives jointly instead of using ensemble-based methods in the future.
%and to focus on consistent spatio-temporal features across the stacked GCNs

% Use \bibliography{yourbibfile} instead or the References section will not appear in your paper
%\nobibliography{aaai22}

%\bibliography{aaai22}

% \section{Acknowledgments}

\bigskip

\end{document}